\documentclass[10pt,twocolumn,letterpaper]{article}

\usepackage{iccv}
\usepackage{times}
\usepackage{epsfig}
\usepackage{graphicx}
\usepackage{amsmath}
\usepackage{amssymb}
\usepackage{multirow}
\usepackage{array}
\usepackage{booktabs} 

\usepackage{blindtext}
\usepackage{caption}

\newcommand\degree{^\circ}

\usepackage[pagebackref=true,breaklinks=true,letterpaper=true,colorlinks,bookmarks=false]{hyperref}

\iccvfinalcopy 

\def\iccvPaperID{****} 

\ificcvfinal\fi

\begin{document}

\title{Conformer: Local Features Coupling Global Representations \\for Visual Recognition}

\author{
Zhiliang Peng$^1$
\quad
Wei Huang$^1$
\quad
Shanzhi Gu$^3$
\quad
Lingxi Xie$^2$
\quad
Yaowei Wang$^3$
\\
Jianbin Jiao$^1$
\quad
Qixiang Ye$^{1,3}$
\\
$^1$University of Chinese Academy of Sciences, Beijing, China
\quad
$^2$Huawei Inc.
\\
$^3$Peng Cheng Laboratory, Shenzhen, China
\\
{\tt\small \{pengzhiliang19, huangwei19\}@mails.ucas.ac.cn \quad \{gushzh, wangyw\}@pcl.ac.cn} \\
\tt\small 198808xc@gmail.com \quad \{jiaojb, qxye\}@ucas.ac.cn
}

\twocolumn[{%
	\maketitle
	\vspace{-0.6cm}
	\renewcommand\twocolumn[1][]{#1}%
	\begin{center}
		\centering
		\includegraphics[width=1.0\textwidth]{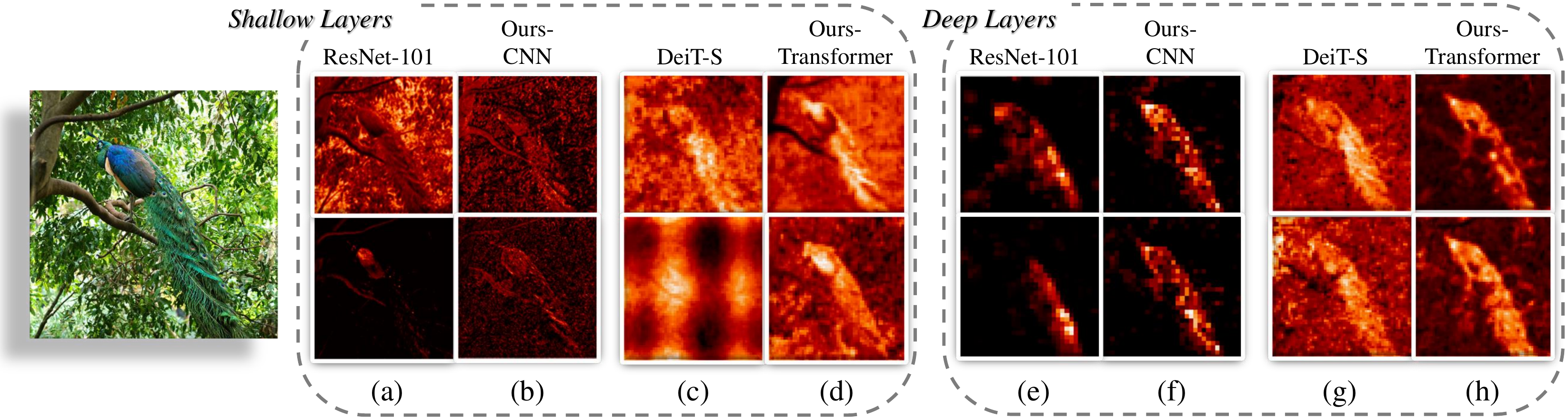}
		\captionof{figure}{
		Comparison of feature maps of CNN (ResNet-101)~\cite{ResNet2016}, Visual Transformer (DeiT-S)~\cite{DeiT2020}, and the proposed Conformer. The patch embeddings in transformer are reshaped to feature maps for visualization. While CNN activates discriminative local regions ($e.g.$, the peacock's head in (a) and tail in (e)), the CNN branch of Conformer takes advantage of global cues from the visual transformer and thereby activates complete object ($e.g.$, full extent of the peacock in (b) and (f)). Compared with CNN, local feature details of the visual transformer are deteriorated ($e.g.$, (c) and (g)). In contrast, the transformer branch of Conformer retains the local feature details from CNN while depressing the background ($e.g.$, the peacock contours in (d) and (h) are more complete than those in (c) and (g)). (Best viewed in color)}
		\label{fig:featuremap}
	\end{center}%
}]


\begin{abstract}
Within Convolutional Neural Network (CNN), the convolution operations are good at extracting local features but experience difficulty to capture global representations.
Within visual transformer, the cascaded self-attention modules can capture long-distance feature dependencies but unfortunately deteriorate local feature details. 
In this paper, we propose a hybrid network structure, termed Conformer, to take advantage of convolutional operations and self-attention mechanisms for enhanced representation learning. 
Conformer roots in the Feature Coupling Unit (FCU), which fuses local features and global representations under different resolutions in an interactive fashion.
Conformer adopts a concurrent structure so that local features and global representations are retained to the maximum extent.
Experiments show that Conformer, under the comparable parameter complexity, outperforms the visual transformer (DeiT-B) by 2.3\% on ImageNet.
On MSCOCO, it outperforms ResNet-101 by 3.7\% and 3.6\% mAPs for object detection and instance segmentation, respectively, demonstrating the great potential to be a general backbone network. 
Code is available at \href{https://github.com/pengzhiliang/Conformer}{\color{magenta}github.com/pengzhiliang/Conformer}. 
\end{abstract}

\section{Introduction}

Convolutional neural networks (CNNs)~\cite{AlexNet2012, VGG2014, GoogLeNet2015, ResNet2016, ResNeXt2017,SENet2018} have significantly advanced computer vision tasks such as image classification, object detection, and instance segmentation. This largely attributes to the convolution operation, which collects local features in a hierarchical fashion as powerful image representations. Despite of the advantage upon local feature extraction, CNNs experience difficulty to capture global representations, $e.g.,$ long-distance relationships among visual elements, which are often critical for high-level computer visual tasks. An intuitive solution is enlarging the receptive field, which however could require more intensive yet damaging pooling operations.

Recently, the transformer architecture~\cite{Transformer2017} has been introduced to visual tasks~\cite{ViT2020, VT2020,DeiT2020,T2TViT2021,DETR2020,PiT2020,ViTFRCNN2020,SETR2020,TransGAN2021}. The ViT method~\cite{ViT2020} constructs a sequence of tokens by splitting each image to patches with positional embeddings and applies cascaded transformer blocks to extract parameterized vectors as visual representations. Thanks to the self-attention mechanism and Multilayer Perceptron (MLP) structure, the visual transformer reflects complex spatial transforms and long-distance feature dependencies, which constitute global representations. Unfortunately, visual transformers are observed ignoring local feature details which decreases the discriminability between background and foreground, Figs.~\ref{fig:featuremap}(c) and (g). Improved visual transformers~\cite{ViT2020,T2TViT2021} have proposed a tokenization module or leveraged CNN feature maps as input tokens to capture feature neighboring information. Nevertheless, the problem about how to precisely embed local features and global representations to each other remains.

In this paper, we propose a dual network structure, termed Conformer, with the aim to couple CNN-based local features with transformer-based global representations for enhanced representation learning. 
Conformer consists of a CNN branch and a transformer branch which respectively follow the design of ResNet~\cite{ResNet2016} and ViT~\cite{ViT2020}.
The two branches form a comprehensive combination of local convolution blocks, self-attention modules, and MLP units.
During training, the cross entropy losses are used to supervise both the CNN and transformer branches to couple CNN-style and transformer-style features.

Considering the feature misalignment between CNN and transformer features, the Feature Coupling Unit (FCU) is designed as the bridge.
On the one hand, to fuse the two-style features, FCU leverages 1$\times$1 convolution to align the channel dimensions, down/up sampling strategies to align feature resolutions, LayerNorm~\cite{LN2016} and BatchNorm~\cite{BN2015} to align feature values. On the other hand, since CNN and transformer branches tend to capture features of different levels ($e.g.$, local vs. global), FCU is inserted into every block to consecutively eliminate the semantic divergence between them, in an interactive fashion. Such a fusion procedure can greatly enhance the global perception capability of local features and the local details of global representations.

The ability of Conformer in coupling local features and global representations is demonstrated in Fig.~\ref{fig:featuremap}. While conventional CNNs ($e.g.$, ResNet-101) tend to retain discriminative local regions ($e.g.$, the peacock's head or tail), the CNN branch of Conformer can activate the full object extent, Figs.~\ref{fig:featuremap}(b) and (f). When solely using the visual transformers, for the weak local features ($e.g.$, blurred object boundaries), it is difficult to distinguish the object from the background, Figs.~\ref{fig:featuremap}(c) and (g). The coupling of local features and global representations significantly enhances the discriminability of transformer-based features, Figs.~\ref{fig:featuremap}(d) and (h).

The contributions of this paper include:
\begin{itemize}
    \item {We propose a dual network structure, termed Conformer, which retains local features and global representations to the maximum extent. }
    
    \item {We propose the Feature Coupling Unit (FCU), to fuse convolutional local features with transformer-based global representations in an interactive fashion. }

    \item Under comparable parameter complexity, Conformer outperforms CNNs and visual transformers by significant margins. Conformer inherits the structure and generalization advantages of both CNNs and visual transformers, demonstrating the great potential to be a general backbone network.
    
\end{itemize}

\begin{figure*}[!ht]
\begin{center}
\includegraphics[width=1\linewidth]{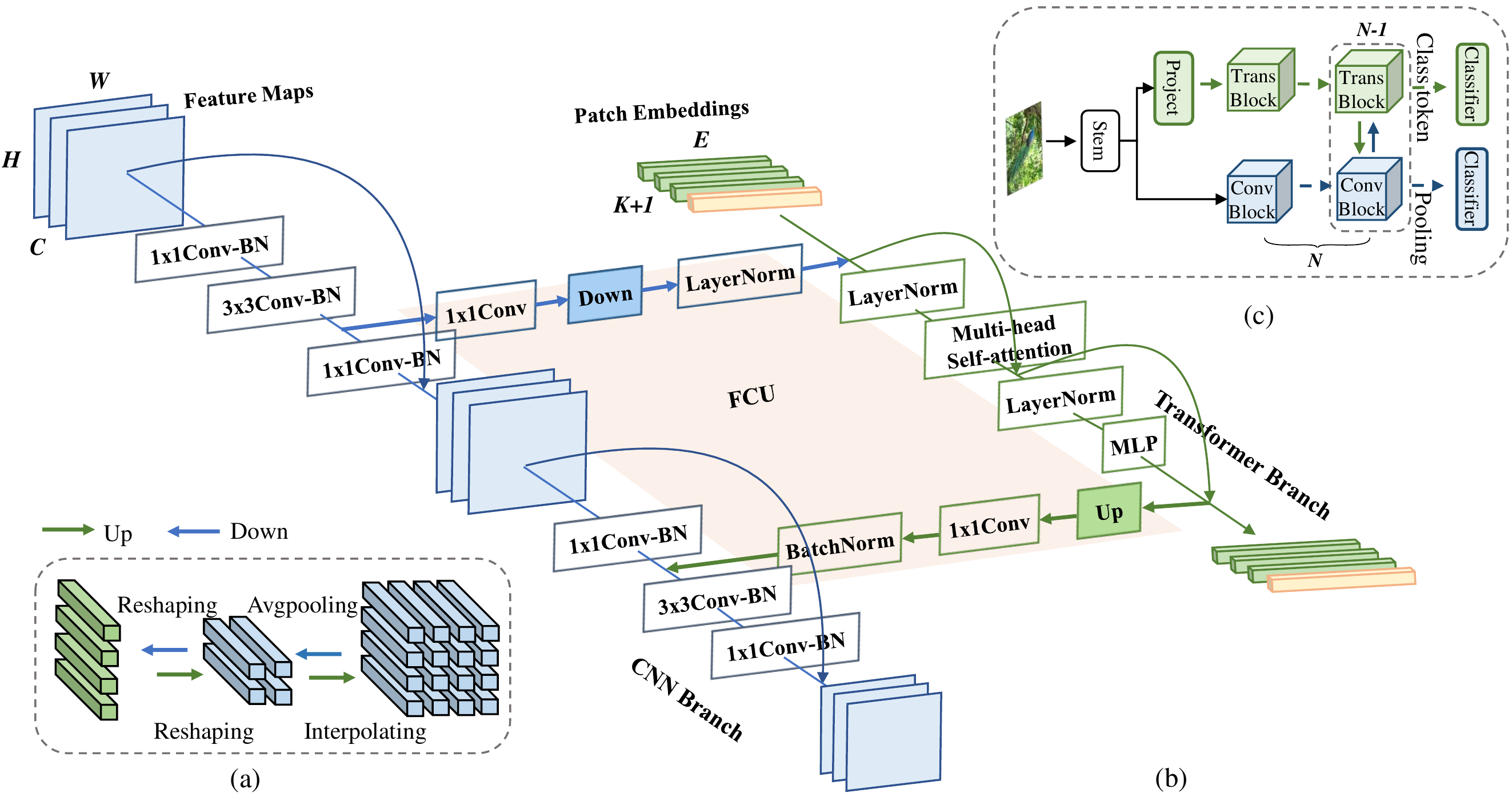}
\end{center}
\vspace{-1em}
\caption{Network architecture of the proposed  Conformer. (a) Up-sampling and down-sampling for spatial alignment of feature maps and patch embeddings. (b) Implementation details of the CNN block, the transformer block, and the Feature Coupling Unit (FCU). (c) Thumbnail of Conformer.}
\label{fig:net}
\end{figure*}

\section{Related Work}

\noindent\textbf{CNNs with Global Cues.}
In the deep learning era, CNNs can be regarded as a hierarchical ensemble of local features with different reception fields. Unfortunately, most CNNs~\cite{AlexNet2012,VGG2014,ResNet2016,Inception2017,ResNeXt2017,DenseNet2017,HRNet2020} are good at extracting local features but experience difficulty to capture global cues. 

To alleviate such a limitation, one solution is to define larger receptive fields by introducing deeper architectures and/or more pooling operations~\cite{SENet2018,Gather-excite2018}. The dilated convolution methods~\cite{DilatedConv2015,Dilated2017} increased the sampling step size, while deformable convolution~\cite{DCN2017} learned the sampling positions. 
SENet~\cite{SENet2018} and GENet~\cite{Gather-excite2018} proposed to use global Avgpooling to aggregate global context and then used it to reweight feature channels, while CBAM~\cite{Cbam2018} respectively used global Maxpooling and global Avgpooling to refine features independently in the spatial and channel dimensions.

The other solution is the global attention mechanism~\cite{NonLocal2018,GCNet2019,AACN2019,RelationNet2018,BoT2021}, which has demonstrated great advantage in capturing long-distance dependencies in natural language processing~\cite{Transformer2017,Bert2018,GPT-3-2020}. Inspired by the non-local means method~\cite{NonLocal2005}, the non-local operation~\cite{NonLocal2018} was introduced to CNNs in a self-attention manner so that the response at each position is a weighted sum of the features at all (global) positions.
Attention augmented convolutional networks~\cite{AACN2019} concatenated convolutional feature maps with self-attentional feature maps to augment convolution operations for capturing long-range interactions.
Relation Networks~\cite{RelationNet2018} proposed an object attention module, which processes a set of objects simultaneously through interaction between their appearance feature and geometry.

Despite of the progress, existing solutions that introduce global cues to CNNs have obvious disadvantages. For the first solution, larger receptive fields require more intensive pooling operations, which implies lower spatial resolution. For the second solution, if convolutional operations are not properly fused with attention mechanisms, local feature details could deteriorate.
~\\

\noindent\textbf{Visual Transformers.}
As a pioneered work, ViT~\cite{ViT2020} validated the feasibility of pure transformer architectures for computer vision tasks. To leverage the long-distance dependencies,  transformer blocks acted as independent architectures or  were introduced to CNNs for image classification~\cite{VT2020,DeiT2020,T2TViT2021}, object detection~\cite{DETR2020,DefomableDETR2020,ViTFRCNN2020}, semantic segmentation~\cite{SETR2020}, image enhancement~\cite{PiT2020} and image generation~\cite{iGPT2020,TransGAN2021}. However, the self-attention mechanism in visual transformers often ignores local feature details. To solve, DeiT~\cite{DeiT2020} proposed using a distillation token to transfer CNN-based features to visual transformer while T2T-ViT~\cite{T2TViT2021} proposed using a tokenization module to recursively reorganize the image to tokens considering neighboring pixels. The DETR method~\cite{DETR2020,DefomableDETR2020} fed local features extracted by CNN to the transformer encoder-decoder to model the global relationships between features in a serial fashion.

\newcommand{\blockb}[2]{\multirow{3}{*}{
\(\left[
\begin{array}{l}
\text{1$\times$1, #2}\\
[-.1em] \text{3$\times$3, #2}\\
[-.1em] \text{1$\times$1, #1}
\end{array}\right]\)}
}

\newcommand{\fcu}[2]{\multirow{7}{*}{
\(\left[
\begin{array}{l}
\text{1$\times$1, #1}\\
[-.1em] \text{1$\times$1, #2}
\end{array}\right]\)}
}

\newcommand{\blockbx}[2]{\multirow{4}{*}{
\(\left[
\begin{array}{l}
\text{1$\times$1, #2}\\
[-.1em] \text{3$\times$3, #2}\\
[-.1em] \text{1$\times$1, #1}
\end{array}\right]\)}
}

\newcommand{\fcux}[3]{\multirow{2}{*}{
\(\left[
\begin{array}{l}
\text{1$\times$1, #1}\\
[-.1em] \text{1$\times$1, #2}
\end{array}\right]\)}
}

\newcommand{\blocktrans}[0]{\multirow{3}{*}{
\(\left[
\begin{array}{c}
\text{\text{MHSA-6}, 384} \\
[-.1em]\text{1$\times$1, 1536} \\
[-.1em] \text{1$\times$1, 384}
\end{array}\right]\)}
}

\newcommand{\dashline}[0]{
- - - - - - - - - - - - - -
}

\newcolumntype{x}[1]{>\centering p{#1pt}}
\newcommand{\ft}[1]{\fontsize{#1pt}{1em}\selectfont}
\renewcommand\arraystretch{1.2}
\setlength{\tabcolsep}{1.2pt}
\begin{table}[ht]
\begin{center}
\resizebox{\linewidth}{!}{
\begin{tabular}{c|c|c|c|c|c}
\toprule
 stage & output & \textbf{CNN Branch} & \textbf{FCU} & \textbf{Transformer Branch} & \\
\hline

\multirow{2}{*}{\text{c1}} & 112$\times$112 & \multicolumn{3}{c}{7$\times$7, 64, stride 2}  \\
\cline{2-6} 
&  56$\times$56 & \multicolumn{3}{c}{ 3$\times$3 max pooling, stride 2} \\
\hline

\multirow{11}{*}{\text{c2}} & \multirow{11}{*}{$56\times56$,197} &{\blockbx{256}{64}} & \multirow{4}{*}{-}  & {4$\times$4, 384, stride 4} & \multirow{4}{*}{$\times$1} \\ \cline{5-5}
  &  &  &  & \blocktrans &\\
  &  &  &  &  & \\
  &  &  &  &  & \\ \cline{3-6}
  &  &\blockb{256}{64}  &  & & \multirow{7}{*}{$\times$3}\\
  &  &  & $[1\times1, 384]\longrightarrow$ & &\\
  &  &  &  &\blocktrans &\\
  &  &\dashline  &  & &\\
  &  &\blockb{256}{64}  &  & &\\
  &  &  & $\longleftarrow[1\times1, 64]$ & \\
  &  &  &  & \\
\hline

\multirow{7}{*}{\text{c3}} &  \multirow{7}{*}{$28\times28$,197}  & \blockb{512}{128} &  &  & \multirow{7}{*}{$\times$4}\\
  &  &  & $[1\times1, 384]\longrightarrow$ &  &  \\
  &  &  & & \blocktrans &  \\
  &  &\dashline  & & &  \\
  &  &\blockb{512}{128}  &  & &  \\
  &  &  & $\longleftarrow[1\times1, 128]$ & &  \\
  &  &  & &  &  \\
\hline

\multirow{7}{*}{\text{c4}} & \multirow{7}{*}{$14\times14$,197} & \blockb{1024}{256} &  &  &  \multirow{7}{*}{$\times$3}\\
  &  &  & $[1\times1, 384]\longrightarrow$ & &  \\
  &  &  & &\blocktrans &  \\
  &  &\dashline  & & & \\
  &  &\blockb{1024}{256}  & & & \\
  &  &  & $\longleftarrow[1\times1, 256]$ & & \\
  &  &  & & \\  
\hline

\multirow{7}{*}{\text{c5}} & \multirow{7}{*}{$7\times7$,197} & \blockb{1024}{256} & &  & \multirow{7}{*}{$\times$1}\\
  &  &  & $[1\times1, 384]\longrightarrow$ & &\\
  &  &  & &\blocktrans &\\
  &  &\dashline  & & &  \\
  &  &\blockb{1024}{256}  & & &  \\
  &  &  & $\longleftarrow[1\times1, 256]$ & &  \\
  &  &  & & &  \\  
\hline

\multirow{2}{*}{\text{classifier}} & \multirow{2}{*}{$1\times1$, 1} & \text{global pooling} & -& \text{class token}\\
\cline{3-5}
& & {$1\times$1,1000} & -& {$1\times$1,1000} \\
\hline

\multicolumn{2}{c|}{\small Parameters} & \multicolumn{3}{c}{\textbf{37.7 M}} \\
\hline
\multicolumn{2}{c|}{\small MACs} & \multicolumn{3}{c}{\textbf{10.6 G}} \\
\bottomrule
\end{tabular}
}
\end{center}
\vspace{-1em}
\caption{Architecture of Conformer-S, where MHSA-6 denotes the multi-head self-attention with heads 6 in transformer block and the fc layer is viewed as 1$\times$1 convolution here. In FCU column, the arrows represent the flow of feature. And in output column, 56$\times$56,197 respectively mean the size of feature map is 56$\times$56 and the number of embedded patches is 197.}
\label{tab:arch}
\end{table}
\renewcommand\arraystretch{1.0}

Different from existing works, Conformer defines the first concurrent network structure which fuses features in an interactive fashion. Such a structure not only naturally inherits the structure advantages of both CNN and transformers but also retains the representation capability of local features and global representations to the maximum extent.

\section{Conformer}


\subsection{Overview}\label{ssec:overview}

Local features and global representations are important counterparts, which have been extensively studied in the long history of visual descriptors. Local features and their descriptors~\cite{Sift04,Gabor1991,LBP2002}, which are compact vector representations of local image neighborhoods, have been the building blocks of many computer vision algorithms. Global representations include, but not limited to, contour representations, shape descriptors, and object typologies at long-distance~\cite{Local-global2008}. In the deep learning era, CNN collects local features in a hierarchical manner via convolutional operations and retains the local cues as feature maps. Visual transformer is believed to aggregate global representations among the compressed patch embeddings in a soft fashion by the cascaded self-attention modules.

In order to take advantage of local features and global representations, we design a concurrent network structure, as shown in Fig.~\ref{fig:net}(c), termed Conformer. 
Considering the complementarity of the two-style features, within Conformer, we consecutively feed the global context from the transformer branch to feature maps, to reinforce the global perception capability of the CNN branch. Similarly, local features from the CNN branch are progressively fed back to patch embeddings, to enrich the local details of the transformer branch. Such a process constitutes the interaction.

In special, Conformer is composed of a stem module, dual branches, FCUs to bridge dual branches, and two classifiers (a fc layer) for the dual branches. 
The stem module, which is a 7$\times$7 convolution with stride 2 followed by a 3$\times$3 max pooling with stride 2, is used to extract initial local features ($e.g.$, edge and texture information), which are then fed to the dual branches. 
The CNN branch and transformer branch are composed of $N$ ($e.g.$, 12) repeated convolution and transformer blocks, respectively, as described in Tab.~\ref{tab:arch}.
Such a concurrent structure implies that CNN and transformer branch can respectively preserve the local features and global representations to the maximum extent.
FCU is proposed as a bridge module to fuse local features in the CNN branch with global representations in the transformer branch, Fig.~\ref{fig:net}(b). FCU is applied from the second block because the initialized features of the two branches are the same. Along the branches, FCU progressively fuses feature maps and patch embeddings in an interactive fashion. 

Finally, for the CNN branch, all the features are pooled and fed to one classifier. For the transformer branch, the class token is taken out and fed to the other classifier. During training, we use two cross entropy losses to separately supervise the two classifiers. The importance of the loss functions are empirically set to be same. During inference, the outputs of the two classifiers are simply summarized as the prediction results.

\subsection{Network Structure}\label{ssec:structure}

\noindent\textbf{CNN Branch.} 
As shown in Fig.~\ref{fig:net}(b), the CNN branch adopts feature pyramid structure, where the resolution of feature maps decreases with network depth while the channel number increases. We split the whole branch into 4 stages, as described in Tab.~\ref{tab:arch}(CNN Branch). Each stage is composed of multiple convolution blocks and each convolution block contains $n_c$ bottlenecks. Following the definition in ResNet~\cite{ResNet2016}, a bottleneck contains a 1$\times$1 down-projection convolution, a 3$\times$3 spatial convolution, a 1$\times$1 up-projection convolution, and a residual connection between the input and output of the bottleneck. In experiments, $n_c$ is set to be 1 in the first convolution block and satisfies $\ge 2$ in the subsequent $N-1$ convolution blocks.

Visual transformers~\cite{ViT2020,DeiT2020} project  an image patch into a vector through a single step, causing the lost of local details. While in CNNs, convolution kernels slide over feature maps with overlap, which provides the possibility to preserve fine-detailed local features. Consequently, the CNN branch is able to consecutively provide local feature details for the transformer branch.
~\\

\noindent\textbf{Transformer Branch.}
Following ViT~\cite{ViT2020}, this branch contains $N$ repeated transformer blocks. As shown in Fig.~\ref{fig:net}(b), each transformer block consists of a multi-head self-attention module and an MLP block (contains a up-projection fc layer and a down-projection fc layer). LayerNorms~\cite{LN2016} are applied before each layer and residual connections in both the self-attention layer and MLP block. For tokenization, we compress the feature maps generated by the stem module into 14$\times$14 patch embeddings without overlap, by a linear projection layer, which is a 4$\times$4 convolution with stride 4. A class token is then pretended to the patch embeddings for classification. Considering that the CNN branch (3$\times$3 convolution) encodes both local features and  spatial location information~\cite{Position2021}, the positional embeddings are no longer required. This facilities increasing image resolution for downstream vision tasks.
~\\

\noindent\textbf{Feature Coupling Unit.}
Given the feature maps in the CNN branch and patch embeddings in the transformer branch, how to eliminate the misalignment between them is an important issue. To solve, we propose the FCU to consecutively couple local features with global representations in an interactive manner.


\begin{figure}[t]
\begin{center}
\includegraphics[width=1.0\linewidth]{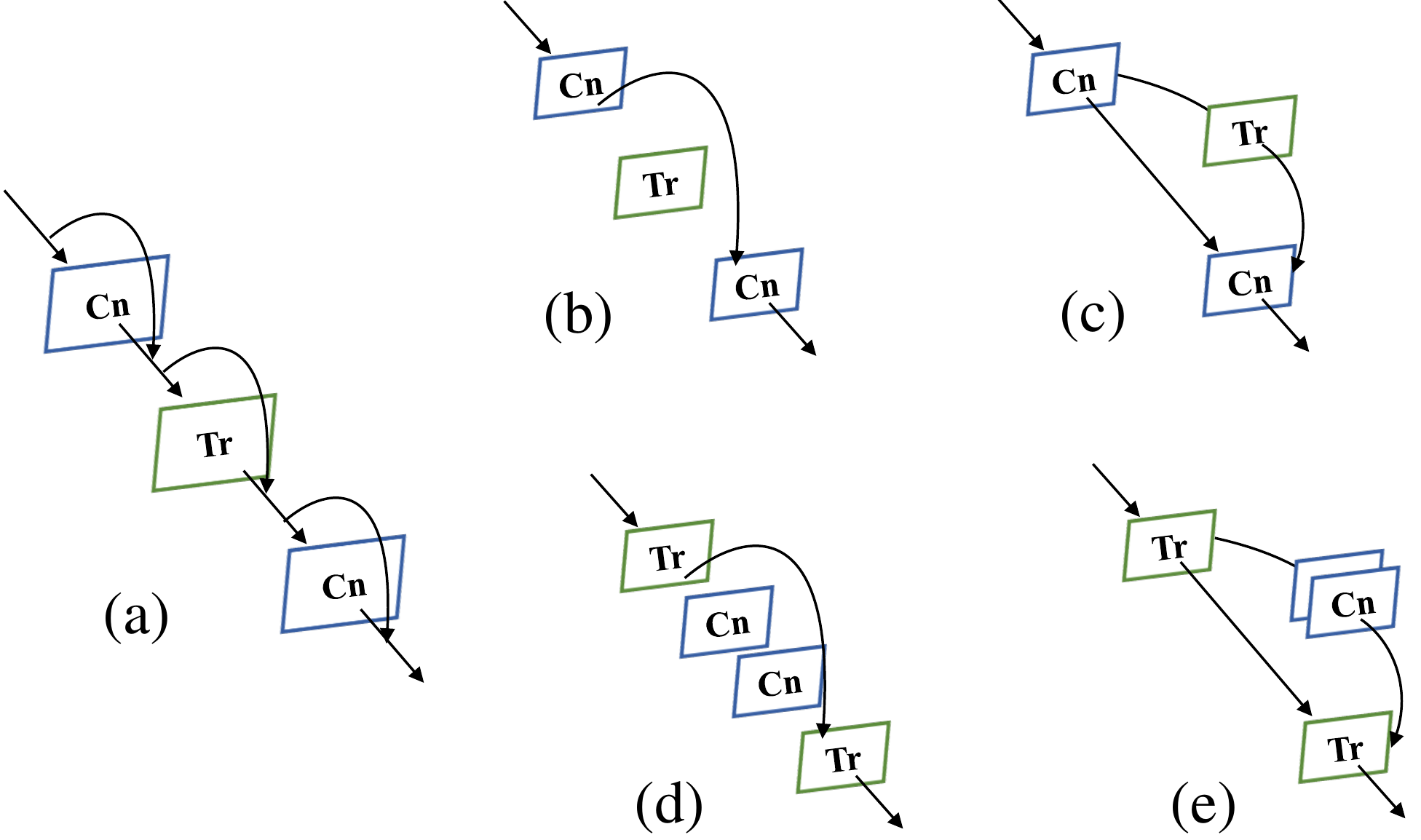}
\end{center}
\vspace{-1em}
\caption{Structure analysis. $C_n$ and $T_r$ respectively denote a bottleneck and a transformer block. (a) The dual structure can be considered as a special serial case of the residual structure. (b) The CNN ($e.g.$, ResNet); (c) A special hybrid structure where the transformer block is embedded to bottlenecks. (d) The visual transformers ($e.g.$, ViT); (e) A special case where the bottlenecks are embedded to the transformer blocks. 
}
\label{fig:simnet}
\end{figure}

%
On the one hand, we must realize that the feature dimensinalities of CNN and transformer are inconsistent. The CNN feature maps have the dimensinality $C\times H\times W$ ($C$, $H$, $W$ are channels, height and width respectively), while the shape of the patch embeddings is $(K+1)\times E$, where $K$, 1, and $E$ respectively represent the number of image patches, class token and embedding dimensions.
When fed to the transformer branch, feature maps first require to get through 1$\times$1 convolution to align the channel numbers of the patch embeddings. A down-sampling module (Fig.~\ref{fig:net}(a)) is then used to complete the spatial dimension alignment. Finally, the feature maps are added with patch embeddings, as shown in Fig.~\ref{fig:net}(b). 
When fed back from the transformer branch to the CNN branch, the patch embeddings require to be up-sampled (Fig.~\ref{fig:net}(a)) to align the spatial scale. The channel dimension is then aligned with that of CNN feature maps through the 1$\times$1 convolution, and added to the feature maps. Meanwhile, LayerNorm and BatchNorm modules are used to regularize features.

On the other hand, there is a significant semantic gap between feature maps and patch embeddings, $i.e.$, feature maps are collected from the local convolutional operators while patch embeddings are aggregated with the global self-attention mechanisms. FCU is therefore applied in each block (except the first) to progressively fill the semantic gap.

\subsection{Analysis and Discussion}\label{ssec:analysis}

\begin{figure}[t]
\begin{center}
\includegraphics[width=\linewidth]{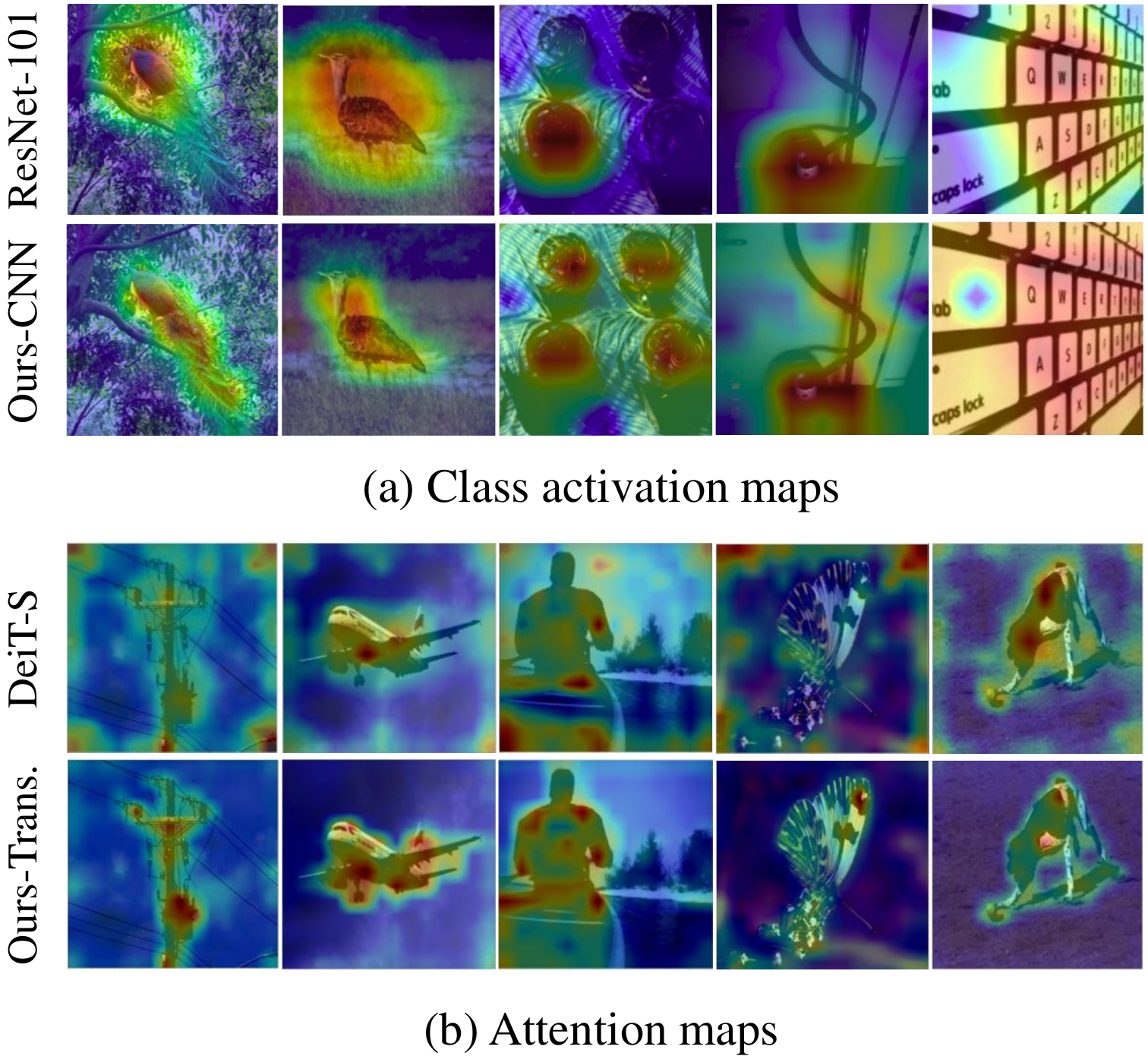}
\end{center}
\vspace{-1em}
\caption{Feature analysis. (a) Class activation maps in ResNet-101 and the CNN branch of Conformer-S by using the CAM method~\cite{CAM2015}. (b) Attention maps in DeiT-S and the transformer branch of Conformer-S by using the Attention Rollout method~\cite{AttentionMap2020}. (Best viewed in color)}
\label{fig:attention}
\end{figure}

\noindent\textbf{Structure Analysis.} 
By considering the FCU as a short connection, we can abstract the proposed dual structure into the special serial residual structure, as shown in Fig.~\ref{fig:simnet}(a). Under different residual connection units, Conformer can implement different depths combinations of bottlenecks (as in ResNet, Fig.~\ref{fig:simnet}(b)) and transformer blocks (as in ViT, Fig.~\ref{fig:simnet}(d)), implying that Conformer inherits the structural advantages of both CNNs and visual transformers. Furthermore, it achieves different permutations of bottlenecks and transformer blocks at different depths, including but not limited to Figs.~\ref{fig:simnet}(c) and (e). This greatly enhances the representation capacity of the network. 
~\\

\noindent\textbf{Feature Analysis.} 
We visualize the feature maps in Fig.~\ref{fig:featuremap}, class activation maps and attention maps in Fig.~\ref{fig:attention}.
Compared with ResNet~\cite{ResNet2016}, with the coupled global representations, the CNN branch of Conformer tends to activate larger regions rather than local areas, suggesting enhanced long-distance feature dependencies, which are significantly demonstrated in Figs.~\ref{fig:featuremap}(f) and~\ref{fig:attention}(a). Thanks to the fine-detailed local features progressively provided by the CNN branch, the patch embeddings of the transformer branch in the Conformer retain important detailed local features (Figs.~\ref{fig:featuremap}(d) and (h)), which are deteriorated by the visual transformers~\cite{ViT2020,DeiT2020} (Figs.~\ref{fig:featuremap}(c) and (g)). Furthermore, the attention area in Fig.~\ref{fig:attention}(b) is more complete while the background is significantly suppressed, implying the higher discriminative capacity of the learned feature representations by Conformer.

\section{Experiments}
\begin{table}[t]
\begin{center}
\begin{tabular}{l|c|cc|c }
\toprule
\multirow{2}{*}{Model}  & Image &   \#Params   &   MACs   &  Top-1  \\
&  size& (M) & (G) & (\%) \\
 \midrule
ResNet-50~\cite{ResNet2016} & $\text{224}^2$ & 25.6 & 4.1  & 76.2 \\
ResNet-101~\cite{ResNet2016} & $\text{224}^2$ & 44.5 & 7.8 &  77.4 \\
ResNet-152~\cite{ResNet2016} & $\text{224}^2$ & 60.2 & 11.6& 78.3 \\
RegNetY-4.0GF~\cite{RegNet2020} & $\text{224}^2$ & 20.6 & 4.0 & 78.8 \\
RegNetY-12.0GF~\cite{RegNet2020} & $\text{224}^2$ & 51.8 & 12.1 & 80.3 \\
RegNetY-32.0GF~\cite{RegNet2020} & $\text{224}^2$ & 145.0 & 32.3 & 81.0 \\
\midrule
ViT-B~\cite{ViT2020} & $\text{384}^2$ & 86 & 55.5 & 77.9 \\
ViT-L~\cite{ViT2020} & $\text{384}^2$ & 307 & 191.1 & 76.5 \\
T2T-$\text{ViT}_{t}$-14~\cite{T2TViT2021} & $\text{224}^2$ & 21.5 & 5.2& 80.7\\
T2T-$\text{ViT}_{t}$-19~\cite{T2TViT2021} & $\text{224}^2$ & 39.0 & 8.4& 81.4\\
T2T-$\text{ViT}_{t}$-24~\cite{T2TViT2021} & $\text{224}^2$ & 64.1 & 13.2& 82.2\\ 
DeiT-S~\cite{DeiT2020} & $\text{224}^2$ & 22.1 & 4.6& 79.8 \\
DeiT-B~\cite{DeiT2020} & $\text{224}^2$ & 86.6 & 17.6&  81.8 \\ 
\midrule
Conformer-Ti & $\text{224}^2$ & 23.5 & 5.2 & 81.3\\
Conformer-S & $\text{224}^2$ & 37.7 & 10.6 & 83.4\\
Conformer-B & $\text{224}^2$ &  83.3 &  23.3 & 84.1 \\

\bottomrule
\end{tabular}
\end{center}
\vspace{-1em}
\caption{Top-1 accuracy for image classification on the ImageNet validation set.}
\label{tab:classification}
\end{table}

\begin{table*}[t]
\begin{center}
\resizebox{\linewidth}{!}{
\begin{tabular}{c | c | c | cc |c c c  c | c c c  c }
\toprule
Method & Backbone & Input size& \#Params & GFLOPs & $\mathrm{AP}^\mathrm{bbox}$ & $\mathrm{AP}^\mathrm{bbox}_\mathrm{S}$ & $\mathrm{AP}^{bbox}_{M}$ & $\mathrm{AP}^\mathrm{bbox}_\mathrm{L}$ & $\mathrm{AP}^\mathrm{segm}$ & $\mathrm{AP}^\mathrm{segm}_\mathrm{S}$ & $\mathrm{AP}^\mathrm{segm}_\mathrm{M}$ & $\mathrm{AP}^\mathrm{segm}_\mathrm{L}$  \\ \midrule
\multirow{4}{*}{FPN} & $\text{ResNet-50}^{\dagger}$~\cite{FPN2017} & (1333, 800) & 41.5 M & 215.8 & 37.4 & 21.2 & 41.0 & 48.1 &  - & - & - & -\\
 & $\text{ResNet-101}^{\dagger}$~\cite{FPN2017} & (1333, 800) & 60.5 M & 295.7 & 39.4 & 22.4 & 43.7 & 51.1 &  - & - & - & - \\
 & Conformer-S/32 & (1344, 800) & 55.4 M & 288.4 & 43.1 & 26.8 &  46.5 & 55.8 & - & - & - & - \\
 & Conformer-S/16 & (1120, 800) & 54.2 M & 404.6 & 44.2 & 28.5 & 48.1 & 58.4 &  - & - & - & - \\
\midrule
\multirow{4}{*}{Mask R-CNN} & $\text{ResNet-50}^{\dagger}$~\cite{MaskRCNN2017} & (1333, 800) & 44.2 M & 268.9 & 38.2 & 21.9 & 40.9 & 49.5 & 34.7 & 18.3 & 37.4 & 47.2 \\
 & $\text{ResNet-101}^{\dagger}$~\cite{MaskRCNN2017} & (1333, 800) & 63.2 M& 348.8 & 40.0 & 22.6 & 44.0 & 52.6 & 36.1 & 18.8 & 39.7 & 49.5 \\
 & $\text{Conformer-S/32}$  & (1344, 800) & 58.1 M & 341.4 & 43.6 & 27.5 &  46.9 & 56.5 & 39.7 & 23.5 & 42.8 & 53.2   \\
 & Conformer-S/16  & (1120, 800) & 56.9 M & 457.7 & 44.9 & 28.7 & 48.8 & 58.6  & 40.7 & 24.4 & 44.3 & 55.1  \\
\bottomrule
\end{tabular}
}
\end{center}
\vspace{-1em}
\caption{Performance for object detection and instance segmentation on the MSCOCO minival set. $\dagger$ means the results are reported by the mmdetection library~\cite{mmdetection}.}
\label{tab:detection}
\end{table*}

\subsection{Model Variants}
By tuning the parameters of the CNN and transformer branches, we have the model variants, termed Conformer-Ti, -S, and -B, respectively. The details of Conformer-S are described in Tab.~\ref{tab:arch}, and those of Conformer-Ti/B are in the Appendix. Conformer-S/32 splits the feature maps to 7$\times$7 patches, $i.e.$, the patch size is 32$\times$32 in the transformer branch.

\subsection{Image Classification}

\noindent\textbf{Experimental Setting.} Conformer is trained on the ImageNet-1k~\cite{ImageNet2009} training set with 1.3M images and tested upon the validation set. The Top-1 accuracy is reported in Tab.~\ref{tab:classification}.
To make the transformer converge to a reasonable performance, we follow the data augmentation and regularization techniques in DeiT~\cite{DeiT2020}. 
These techniques include Mixup~\cite{Mixup2017}, CutMix~\cite{CutMix2019}, Erasing~\cite{Erasing2020}, Rand-Augment~\cite{RA2020} and Stochastic Depth~\cite{StochasticDepth2016}). The model is trained for 300 epochs with the AdamW optimizer~\cite{AdamW2017}, batchsize 1024 and weight decay 0.05. 
The initial learning rate is set to 0.001 and decay in a cosine schedule.
~\\

\noindent\textbf{Performance.} Under similar parameters and computational budgets, Tab.~\ref{tab:classification}, Conformers outperform both CNN and visual transformers. For example, Conformer-S (with 37.7M parameters and 10.6G MACs) respectively outperforms ResNet-152 (with 60.2M parameters and 11.6G MACs) by 4.1\%((83.4\% vs. 78.3\%) and DeiT-B (with 86.6M parameters and 17.6G MACs) by 1.6\% (83.4\% vs. 81.8\%). Conformer-B, with comparable parameters and moderate MAC cost, outperforms DeiT-B by 2.3\% (84.1\% vs. 81.8\%). Beyond its superior performance, Conformer converges faster than the visual transformers.

\subsection{Object Detection and Instance Segmentation}
To verify Conformer's versatility, we test it on instance-level tasks ($e.g.$, object detection) and pixel-level tasks ($e.g.$, instance segmentation) on the MSCOCO dataset\footnote{Using mmdetection library at \href{github.com/open-mmlab/mmdetection}{\color{magenta}github.com/open-mmlab/mmdetection}}~\cite{MSCOCO2014}. Conformer, as the backbone, is migrated without extra design, and the relative accuracy and parameter comparison is included in Tab.~\ref{tab:classification}. With the CNN branch, we can use the output feature maps of $[\mathrm{c}_\mathrm{2}, \mathrm{c}_\mathrm{3}, \mathrm{c}_\mathrm{4}, \mathrm{c}_\mathrm{5}]$ as side-output to construct the feature pyramid~\cite{FPN2017}.
~\\

\noindent\textbf{Experimental Setting.}
As is common practice, the models are trained on the MSCOCO training set and tested on the MSCOCO minival set. In Tab.~\ref{tab:detection}, we report $\mathrm{AP}^\mathrm{bbox}$ ($\mathrm{AP}^\mathrm{segm}$), $\mathrm{AP}^\mathrm{bbox}_{S}$ ($\mathrm{AP}^\mathrm{segm}_{S}$), $\mathrm{AP}^\mathrm{bbox}_{M}$ ($\mathrm{AP}^\mathrm{segm}_{M}$), and $\mathrm{AP}^\mathrm{bbox}_{L}$ ($\mathrm{AP}^\mathrm{segm}_{L}$) for averaged over IoU thresholds, small, medium and large objects of box (mask), respectively. Unless explicitly specified, we use the batch size 32, with a learning rate 0.0002, optimizer AdamW~\cite{AdamW2017}, weight decay 0.0001 and max epoch 12. The learning rate decays at the 8-$\mathrm{th}$ and 11-$\mathrm{th}$ epoch by a magnitude.
~\\

\begin{table}[t]
\centering
\begin{tabular}{c|c|c|c|c|c|c|c|c}
\toprule
\multicolumn{3}{c|}{\small Transformer branch}& \multicolumn{3}{c|}{\small CNN branch}      & \multirow{2}{*}{$p_p$} & \multirow{2}{*}{MACs} & \multirow{2}{*}{Acc.(\%)} \\ \cline{1-6}
$E$                    & $d_h$                   &  \#Params             & $n_c$                 & $C$   & \#Params &                         &                          &                          \\ \hline
\multirow{8}{*}{384} & \multirow{8}{*}{6}  & \multirow{8}{*}{22 M}   & -                  & -   & -         & -                       & 4.6 G                      & 79.8                     \\ \cline{4-9} 
                     &                     &                       & \multirow{5}{*}{2} & 64  & 1.5 M       & 0.07                    & 5.2 G                      & 81.3                     \\ \cline{5-9} 
                     &                     &                       &                    & 128 & 4.5 M       & 0.2                     & 6.4 G                      & 82.3                     \\ \cline{5-9} 
                     &                     &                       &                    & 192 & 9.3 M       & 0.4                     & 8.2 G                       & 82.8                     \\ \cline{5-9} 
                     &                     &                       &                    & 256 & 15.7 M      & 0.7                     & 10.6 G                      & 83.4                     \\ \cline{5-9} 
                     &                     &                       &                    & 320 & 23.7 M      & 1.0                     & 13.7 G                     & 83.6                     \\ \cline{4-9} 
                     &                     &                       & 4                  & 192 & 15.8 M      & 0.7                     & 10.9 G                      & 83.3                     \\ \cline{4-9} 
                     &                     &                       & 3                  & 256  &21.4 M   &1.0    & 13.0G       & 83.5                             \\ \hline
\multirow{3}{*}{576} & \multirow{3}{*}{9}  & \multirow{3}{*}{48.9 M} & -                  & -   & -         & -                       & 10.0 G                      & 79.0                     \\ \cline{4-9} 
                     &                     &                       & \multirow{2}{*}{2} & 256 & 16.4 M      & 0.3                     & 16.3 G                      & 83.6                     \\ \cline{5-9} 
                     &                     &                       &                    & 384 & 36.4 M      & 0.7                     & 23.3 G                      & 84.1                     \\ \hline
\multirow{2}{*}{768} & \multirow{2}{*}{12} & \multirow{2}{*}{86 M}   & -                  & -   & -         & -                       & 17.6 G                      & 81.8                     \\ \cline{4-9} 
                     &                     &                       & 2                  & 256 & 17.6 M      & 0.2                     & 24.2 G                     & 83.0                     \\ \bottomrule
\end{tabular}%
\caption{Performance under different parameter proportions. $E$ and $d_h$ respectively denote the embedding dimensions and the head in the multi-head attention module in the transformer branch. $C$ and $n_c$ respectively represent the channels of $\mathrm{c_2}$ and the bottleneck number within each convolution block in the CNN branch. $p_p$ is the proportion of CNN (including stem and FCUs) and transformer branch parameters. 
}
\label{tab:parameters}
\end{table}

\renewcommand\arraystretch{1.0}

\noindent\textbf{Performance.}
As shown in Tab.~\ref{tab:detection}, Conformer significantly boosts the $\mathrm{AP}^\mathrm{bbox}$ and $\mathrm{AP}^\mathrm{segm}$. For object detection, the mAP of Conformer-S/32 (55.4 M \& 288.4 GFLOPs) is 3.7\% higher than that of the FPN baseline (ResNet-101, 60.5 M \& 295.7 GFLOPs). For instance segmentation, the mAP of Conformer-S/32 (58.1M \& 341.4 GFLOPs) is 3.6\% higher than that of the Mask R-CNN baseline (ResNet-101, 63.2 M \& 348.8 GFLOPs). This demonstrates the importance of global representations for high level tasks and suggests the great potential of Conformer to be a general backbone network.

\subsection{Ablation Studies}

\noindent\textbf{Number of Parameters.}
The parameters of the proposed Conformer are combinations of the CNN and transformer branches. The parameter proportion of the two branches is a hyper-parameter to be experimentally determined.
In Tab.~\ref{tab:parameters}, we evaluate performance of the two branches under different parameter settings. For the CNN branch, we tune the parameters of the CNN branch by changing the channels and the number of bottlenecks, which respectively control the width and depth of the CNN branch. For the transformer branch, we tune the parameters by changing the numbers of embedding dimensions and heads. From Tab.~\ref{tab:parameters}, one can see that the accuracy is improved by increasing either parameters of the CNN or the transformer branch. More CNN parameters bring greater improvement while the computational cost overhead is lower. 

%
\noindent\textbf{Dual Structure.}
Conformer is a dual model, which is totally different from the serial hybrid ViT (CNN $\rightarrow$ Transformer)~\cite{ViT2020}. In Tab.~\ref{tab:hybrid}, ResNet-26/50d \& DeiT-S is a hybrid model which consists of ResNet-26/50d~\cite{ResNet2016} and DeiT-S~\cite{DeiT2020}, where DeiT-S forms tokens upon the feature maps extracted by ResNet-26/50d. With comparable computational cost overhead, Conformer-S/32 outperforms the serial hybrid model although ResNet-26/50d can retain more local information within the stem stage.
~\\

\begin{table}[t]
\begin{center}
\begin{tabular}{m{3cm}<{\centering} | m{2cm}<{\centering} m{1.1cm}<{\centering} | m{1.8cm}<{\centering}}
\toprule
Model  & \#Params & MACs & Accuracy \\
\midrule
\small DeiT-S/32 & 22.9 M & 1.1 G & 73.8\% \\
\small ResNet-26d \& DeiT-S  & 36.5 M & 3.7 G & 80.2\%  \\
\small ResNet-50d \& DeiT-S  & \textbf{46.0 M} & 5.5 G & 80.4\% \\
\small Conformer-S/32 & 38.8 M & \textbf{7.0 G} & \textbf{81.9\%} \\
\bottomrule
\end{tabular}
\end{center}
\vspace{-1em}
\caption{Comparison of hybrid structures. DeiT-S/32 means the patch size is 32$\times$32 for the DeiT-S model~\cite{DeiT2020}. ResNet-26/50d is the variant of ResNet-26/50, and its stem module is composed of three 3$\times$3 convolutions.}
\label{tab:hybrid}
\end{table}
\begin{table}[t]
\begin{center}
\begin{tabular}{c|c|c}
\toprule
Method  & Pos. Embeds & Accuracy \\
\midrule
\multirow{2}{*}{Deit-S} & $\surd$& 79.8\% \\
& $\times$& 77.4\% (\textbf{-2.4\%}) \\ \midrule
\multirow{2}{*}{Conformer-S} & $\surd$& 83.5\% \\
&   $\times$& 83.4\% (\textbf{-0.1\%}) \\
\bottomrule
\end{tabular}
\end{center}
\vspace{-1em}
\caption{Comparison of positional embeddings strategies. ``Pos. Embeds" is the abbreviation for ``learnable positional embeddings".}
\label{tab:postition}
\end{table}

\noindent\textbf{Positional Embeddings.}
Considering that the CNN branch (3$\times$3 convolution) encodes both local features and spatial location information, the positional embeddings are assumed no longer required for Conformer. In Tab.~\ref{tab:postition}, when the positional embedding is removed, the accuracy of DeiT-S decreases 2.4\%, while that of Conformer-S decreases marginally (0.1\%).
~\\

\noindent\textbf{Sampling Strategies.}
In FCU, to make CNN-based feature maps coupling with Transformer-based patch embeddings, up/down-sampling operations are used to align the spatial scale. In Tab.~\ref{tab:sampling}, we compare different up/down-sampling strategies including Maxpooling, Avgpooling, convolution and attention-based sampling\footnote{Refer to Appendix for detailed attention-based sampling.}. Compared with Max/Avgpooling sampling, convolution and attention-based sampling methods use more parameters and computation cost but achieve comparable accuracy. We thereby choose the Avgpooling strategy.
~\\

\begin{table}[t]
\begin{center}
\resizebox{0.9\linewidth}{!}{
\begin{tabular}{m{1.8cm}<{\centering} m{1.8cm}<{\centering} | m{1.3cm}<{\centering} m{1cm}<{\centering} | c}
\toprule
Down & Up & \#Params & MACs  & Accuracy \\
\midrule
Maxpooling & Interpolation & 37.7 M & 10.3 G & 83.3\% \\
Avgpooling & Interpolation & 37.7 M & 10.3 G & \textbf{83.4\%} \\
Convolution & Interpolation & \textbf{47.7 M} & \textbf{12.3 G} & \textbf{83.4\%}  \\
Attention & Attention &  39.4 M & 11.3 G & 83.3\%  \\
\bottomrule
\end{tabular}
}
\end{center}
\vspace{-1em}
\caption{Comparison of sampling strategies. The nearest neighbor interpolation is used.}
\label{tab:sampling}
\end{table}
\begin{table}[t]
\begin{center}
\resizebox{\linewidth}{!}{
\begin{tabular}{c | c c | c c | c}
\toprule
Model & \#Params & MACs  & $\text{Acc}^{\small C_n}$ & $\text{Acc}^{\small T_r}$ & $\text{Acc}^{All}$\\
\midrule
DeiT-S & 22.0 M & 4.2 G & - & 79.8\% & 79.8\% \\
ResNet-101 & 44.5 M & 7.8 G & 80.6\% & - & 80.6\% \\
DeiT-S + ResNet-101 & \textbf{66.5 M} & \textbf{11.2 G} & 80.6\% & 79.8\% & 81.8\% \\ \midrule
Conformer-S & 37.7 M & 10.3 G & \textbf{83.3\%} & \textbf{83.1\%} & \textbf{83.4\%} \\
\bottomrule
\end{tabular}
}
\end{center}
\vspace{-1em}
\caption{Performance comparison of ensemble models. $\text{Acc}^{\small C_n}$ and $\text{Acc}^{\small T_r}$ respectively denote the accuracy of the CNN and transformer branches.}
\label{tab:ensemble}
\end{table}

\noindent\textbf{Comparison with Ensemble Models.}
Conformer is compared with the ensemble models combining the outputs of CNN and transformer. For fair comparison, we use the same data augmentation and regularization strategies and the same training epochs (300) to train ResNet-101~\cite{ResNet2016}, and combine it with the DeiT-S~\cite{DeiT2020} model to form an ensemble model, and report the accuracy in Tab.~\ref{tab:ensemble}. The accuracies of the CNN branch, the transformer branch, and the Conformer-S respectively reach 83.3\%, 83.1\%, and 83.4\%. 
In contrast, the ensemble model (DeiT-S+ResNet-101) archives 81.8\%, which is 1.6\% lower than that of Conformer-S (83.4\%), although it uses significantly more parameters and MACs. 

\subsection{Generalization Capability}

\begin{figure}[t]
\begin{center}
\includegraphics[width=\linewidth]{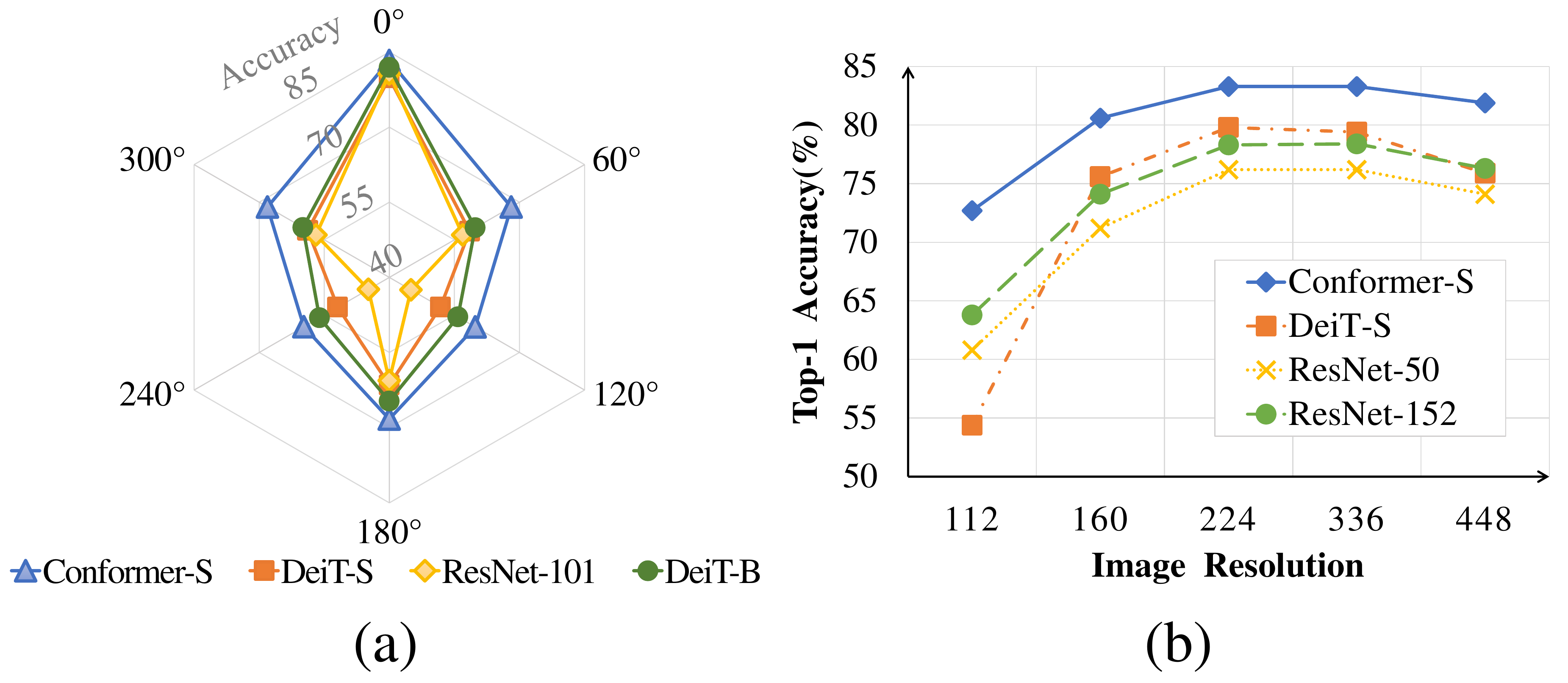}
\end{center}
\vspace{-1em}
\caption{Generalization capability. (a) Comparison of rotation invariance. The compared models are trained under the same data augmentation settings and directly evaluated on rotated images without model fintuning. (b) Comparison of scale invariance. The models are trained on images with the resolution of 224$\times$224, and tested on different image resolutions without model finetuning.}
\label{fig:generalization}
\end{figure}

\noindent\textbf{Rotation Invariance.}
To verify the generalization capability of the model in terms of rotation, we rotate test images by $0\degree$, $60\degree$, $120\degree$, $180\degree$, $240\degree$ and $300\degree$ and evaluate the performance of models trained under same data augmentation settings.  As shown in Fig.~\ref{fig:generalization}(a), all models report comparable performance for images without rotation ($0\degree$). For the rotated test images, the performance of ResNet-101 drops significantly. In contrast, Conformer-S reports higher performance, which implies stronger rotation invariance.
~\\

\noindent\textbf{Scale Invariance.}
In Fig.~\ref{fig:generalization}(b), we compare the scale adaptation ability of Conformer with those of visual transformers (DeiT-S) and CNN (ResNet). 
We interpolate the positional embeddings of DeiT-S to adapt it to input images of different resolutions during inference.
When the size of input images reduces from 224 to 112, DeiT-S's performance drops by 25\% and that of ResNet-50/152 drops by 15\%. In contrast, the performance of Conformer drops only by 10\%, demonstrating higher scale invariance of the learned feature representations.

\section{Conclusion}

We propose Conformer, the first dual backbone to combining CNN with visual transformer. Within Conformer, we leverage the convolution operators to extract local features and the self-attention mechanisms to capture global representations. We design the Feature Coupling Unit (FCU) to fuse local features and global representations, enhancing the ability of visual representations in an interactive fashion. Experiments show that Conformer, with comparable parameters and computation budgets, outperforms both conventional CNNs and visual transformers, in striking contrast with the state-of-the-arts. On downstream tasks, Conformer has shown the great potential to be a simple yet effective backbone network.

{\small
\bibliographystyle{ieee_fullname}
\bibliography{egbib}
}

\appendix
\section{Model Architectures}
The architectures of Conformer-Ti/B are detailed in Tab.~\ref{tab:arch_ti_b}. 
Compared with Conformer-S, Conformer-Ti reduces channel number of the CNN branch by 1/4, and Conformer-B increases channel number in the CNN branch, head number of the multi-head attention module and the embedding dimensions in the transformer branch by 1.5. 

\section{Attention-based Sampling}
We also design a down-sampling-up-sampling strategy based on the cross attention between feature maps and patch embeddings.

Let $h$, $w$ and $c$ respectively denote the height, width, channel of feature maps in a block (we omit the batch dimension here for simplicity), $K$ and $E$ respectively represent the number of patch embeddings (termed $P_t$) and channel dimension in the transformer branch. We split the feature maps into $K$ patches ($e.g.$, 14$\times$14), termed $P_c$. The dimension of each patch is $n\times c$. 
After aligning the channel dimension by 1$\times$1 convolution, the shape of each patch is $n\times E$.

For down sampling, the fusion between patch $i$ in $P_c$ (denoted $P^i_c$) and patch $j$ in $P_t$ (denoted $P^j_t$) is formulated as
\begin{equation}\label{eq:1}
    P^j_t = P^j_t + \operatorname{Softmax}\left(\frac{({P^j_t}{W_q})({P^i_c}{W_k})^T}{\sqrt{E}}\right)({P^i_c}{W_v}),
\end{equation}
where $W_q, W_k, W_v\in \mathbb{R}^{E\times E}$ are learned linear transformations which map the input $P^j_t$ to queries $Q$, keys $K$ and values $V$, respectively. 

For up sampling, we re-use the attention weights in Eq.~\ref{eq:1} and formulate the process as 
\begin{equation}\label{eq:2}
    \Tilde{P}^i_c = \Tilde{P}^i_c + \operatorname{Softmax}\left(\frac{({P^j_t}{W_q})({P^i_c}{W_k})^T}{\sqrt{E}}\right)^T {\Tilde{P}^j_t},
\end{equation}
where $\Tilde{P}^i_c$ and $\Tilde{P}^i_c$ respectively denote that ${P}^i_c$ is processed by convolution layers and ${P}^j_t$ by a transformer block ( Fig. 2 in the paper).

\section{Inference Time}

\begin{figure}[t]
\begin{center}
\includegraphics[width=\linewidth]{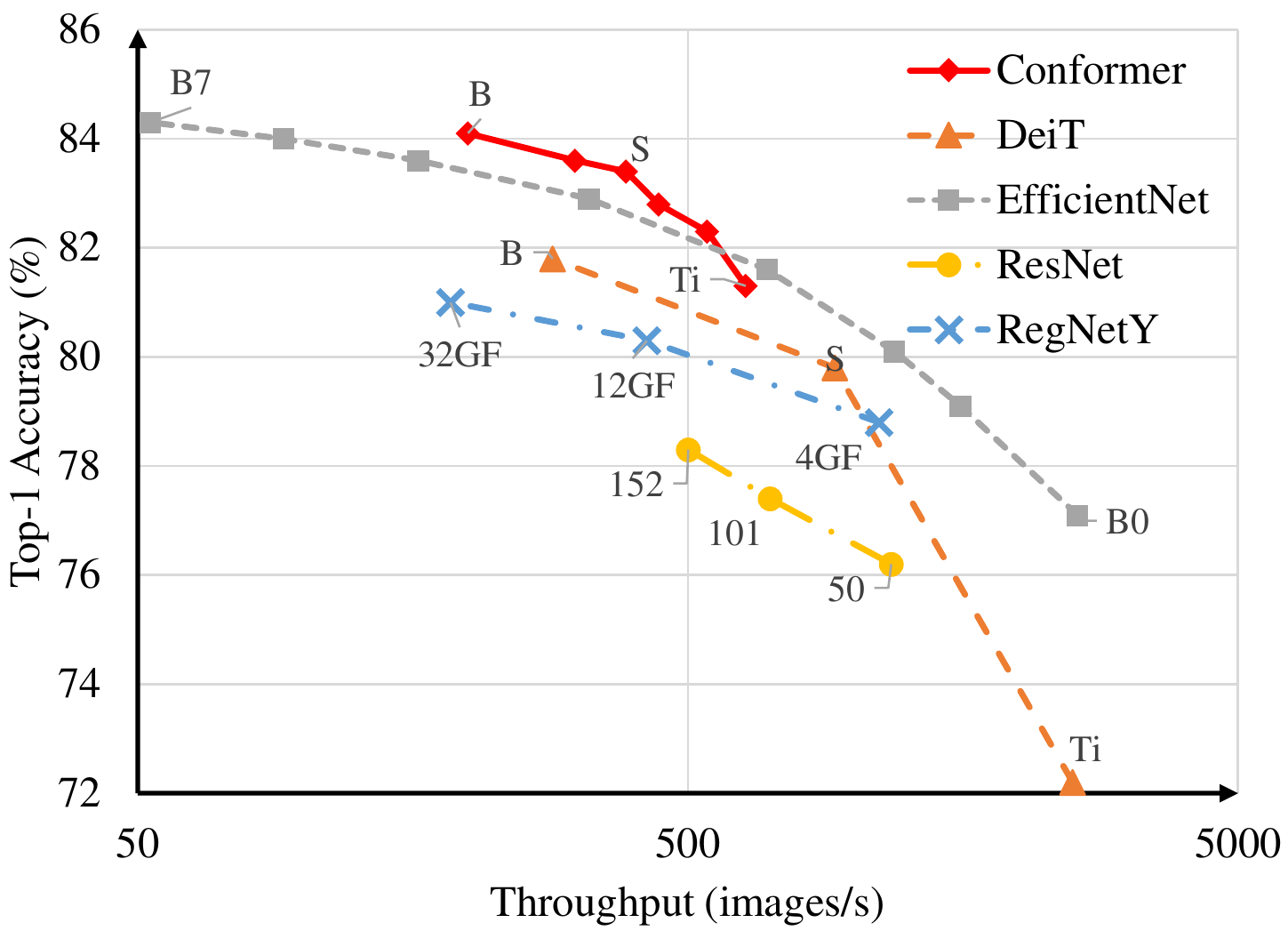}
\end{center}
\vspace{-1em}
\caption{Throughput and accuracy on ImageNet of Conformer compared to DeiT~\cite{DeiT2020}, ResNet~\cite{ResNet2016}, RegNetY~\cite{RegNet2020} and EfficientNet~\cite{Efficientnet2019}. The throughput is measured as the number of images processed per second on a 32GB V100 GPU.
}
\label{fig:throughput}
\end{figure}

\noindent\textbf{Classification.} Following DeiT~\cite{DeiT2020}, we evaluate and compare the throughput of various methods in Fig.~\ref{fig:throughput}. One can see that our Conformer outperforms EfficientNet~\cite{Efficientnet2019} under comparable throughput.
~\\

\noindent\textbf{Object detection and instance segmentation.} 
Similarly, we measure Frame Per Second (FPS) as the inference speed and show the comparison in the Tab.~\ref{tab:fps}. 
Combining Tab.3 in the paper and Tab.~\ref{tab:fps} here, compared with ResNet-101~\cite{ResNet2016}, Conformer-S/32 has the comparable parameters, GFLOPs and inference speed, but can outperform ResNet-101 by a significant margin on both object detection and instance segmentation tasks, which further demonstrates the potential to be a general backbone network.

\begin{table}[t]
    \centering
    \resizebox{1.0\linewidth}{!}{
    \begin{tabular}{c | c|cc|c}
    \toprule
         Method & Backbone & \#Params (M) & GLOPs & FPS  \\ \midrule
         \multirow{4}{*}{FPN} & ResNet-50 & 41.5 & 215.8 & 20.2\\
          & ResNet-101 & 60.5 & 295.7 & 15.9\\
         & Conformer-S/32 & 55.4 & 288.4 & 13.5 \\ 
         & Conformer-S & 54.2 & 404.6 & 8.2 \\ \midrule
         \multirow{4}{*}{Mask R-CNN} & ResNet-50 & 44.2 & 268.9 & 13.2\\
          & ResNet-101 & 63.2 & 348.8 & 11.5 \\
         & Conformer-S/32 & 58.1 & 341.4 & 10.9 \\ 
         & Conformer-S & 56.9 & 457.7 & 7.1 \\
    \bottomrule
    \end{tabular}
    }
    \caption{Comparison of inference time. FPS is measured on a 32GB V100 GPU with batchsize 1.}
    \label{tab:fps}
\end{table}

\section{Residual Structure}
As shown in Fig. 3 in the paper, by considering FCUs as short connection we abstract Conformer with a dual structure to a serial structure with residual connections. 
In other words, under different residual connections, Conformer can degenerate to different sub-structures. We test some sub-structures and report the corresponding performance in Tab.~\ref{tab:substructure}.
From Tab.~\ref{tab:substructure}, one can see that the proposed residual structure outperforms other  sub-structures.
\begin{table}[h]
    \centering
    \resizebox{0.9\linewidth}{!}{
    \begin{tabular}{c|cc|c}
    \toprule
         Index & \#Params (M) & MACs (G) & Accuracy (\%)  \\ \midrule
         1 & 8.6 & 9.2 & 73.9 \\
         2 & 37.0 & 10.8 & 80.8\\
         3 & 22.1 & 4.6 & 79.8 \\
         4 & 28.9 & 6.0 & 80.2 \\ \midrule
         Conformer-S & 37.7 & 10.6 & 83.4 \\
    \bottomrule
    \end{tabular}
    }
    \caption{Performance of Conformer sub-structures. Where the index 1, 2, 3 and 4 respectively represent the sub-structures shown in Figs. 3(b), (c), (d) and (e).}
    \label{tab:substructure}
\end{table}

\section{Fusion Interval}
In the paper, we proposed a Feature Coupling Unit to interact the local features and global representations in each block to progressively align the features to fill the semantic gap.
To validate whether fusion should be done in each block, we conduct experiments on fusion intervals and report the performance on ImageNet in Tab.~\ref{tab:fusionfeq}.
From Tab.~\ref{tab:fusionfeq}, one can see that 
smaller fusion intervals report higher performance, implying that frequent interaction facilities the representation learning.

\begin{table}[h]
    \centering
    \resizebox{0.9\linewidth}{!}{
    \begin{tabular}{c|cc|c}
    \toprule
         Interval & \#Params (M) & MACs (G) & Accuracy (\%)  \\ \midrule
         1 & 37.7 & 10.6 & 83.4 \\
         2 & 34.2 & 9.2 & 82.9 \\
         4 & 32.3 & 8.4 & 82.2 \\ 
    \bottomrule
    \end{tabular}
    }
    \caption{Comparison of fusion intervals. 1, 2 and 4 respectively represent performing fusion every 1, 2 and 4 block(s).}
    \label{tab:fusionfeq}
\end{table}

\section{Convergence speed}

\begin{figure}[!h]
\begin{center}
\includegraphics[width=0.8\linewidth]{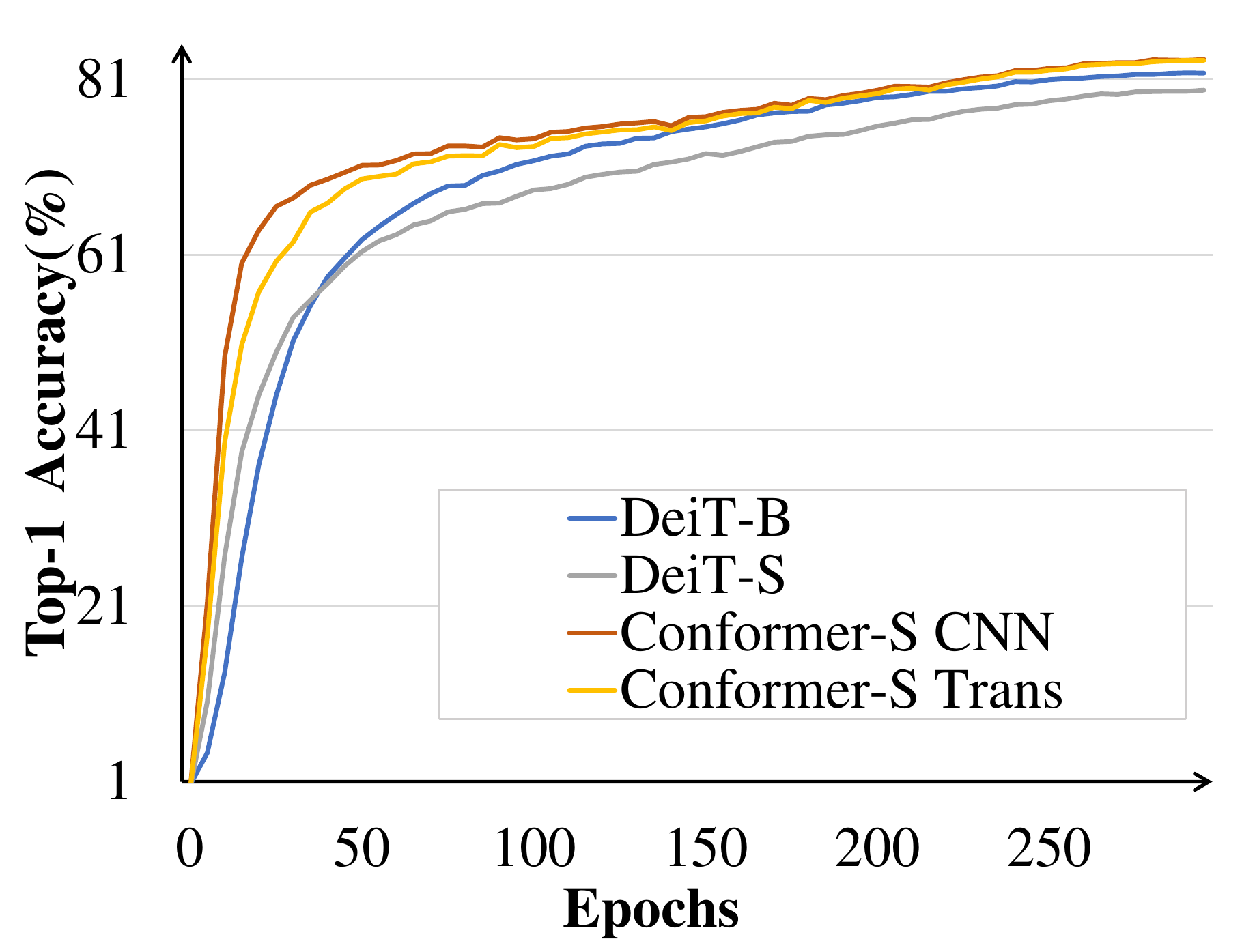}
\end{center}
\vspace{-1em}
\caption{Training Accuracy on the  val set.}
\label{fig:convergence}
\end{figure}

For the convolution operations introduced, Fig.~\ref{fig:convergence}, both the CNN branch and the transformer branch of Conformer-S significantly outperforms DeiT during the first 50 epochs. 
This demonstrates the inductive bias of convolution facilities the convergence of visual transformers.

\newcommand{\blockbc}[3]{\multirow{3}{*}{
\(\left[
\begin{array}{l}
\text{1$\times$1, #2}\\
[-.1em] \text{3$\times$3, #2}\\
[-.1em] \text{1$\times$1, #1}
\end{array}\right]\)}
}

\newcommand{\blockbxc}[3]{\multirow{4}{*}{
\(\left[
\begin{array}{l}
\text{1$\times$1, #2}\\
[-.1em] \text{3$\times$3, #2}\\
[-.1em] \text{1$\times$1, #1}
\end{array}\right]\)}
}

\newcommand{\blocktransc}[1]{\multirow{3}{*}{
\(\left[
\begin{array}{c}
\text{\text{MHSA-6}, 384} \\
[-.1em]\text{1$\times$1, 1536} \\
[-.1em] \text{1$\times$1, 384}
\end{array}\right]\)}
}

\newcommand{\blocktransbc}[1]{\multirow{3}{*}{
\(\left[
\begin{array}{c}
\text{\text{MHSA-9}, 576} \\
[-.1em]\text{1$\times$1, 2304} \\
[-.1em] \text{1$\times$1, 576}
\end{array}\right]\)}
}

\newcommand{\fcuxc}[3]{\multirow{4}{*}{
\(\left[
\begin{array}{l}
\text{1$\times$1, #1}\\
[-.1em] \text{1$\times$1, #2}
\end{array}\right]\)$\times$#3}
}
\newcommand{\fcuc}[3]{\multirow{3}{*}{
\(\left[
\begin{array}{l}
\text{1$\times$1, #1}\\
[-.1em] \text{1$\times$1, #2}
\end{array}\right]\)$\times$#3}
}

\newcommand{\ddashline}[0]{
- - - - - - - - - - - -
}

\newcolumntype{x}[1]{>\centering p{#1pt}}
\newcommand{\ftc}[1]{\fontsize{#1pt}{1em}\selectfont}
\renewcommand\arraystretch{1.6}
\setlength{\tabcolsep}{1.2pt}
\begin{table*}[h]

\begin{center}

\resizebox{\linewidth}{!}{

\begin{tabular}{c|c|c|c|c|c|c|c|c|c|c}

\toprule
& \multicolumn{5}{c|}{ \textbf{ CNNTransformer-Ti}} &\multicolumn{5}{c}{ \textbf{ CNNTransformer-B}} \\
\hline

 stage & output & \textbf{CNN Branch} & \textbf{fcuc} & \textbf{Transformer Branch}& & output & \textbf{CNN Branch}& \textbf{fcuc} & \textbf{Transformer Branch}& \\
\hline
\multirow{2}{*}{\text{c1}} & 112$\times$112 & \multicolumn{4}{c|}{7$\times$7, 64, stride 2}& 112$\times$112 & \multicolumn{4}{c}{7$\times$7, 64, stride 2}  \\
\cline{2-11} 
&  56$\times$56 & \multicolumn{4}{c|}{ 3$\times$3 max pooling, stride 2}&  56$\times$56 & \multicolumn{4}{c}{ 3$\times$3 max pooling, stride 2} \\
\hline

\multirow{11}{*}{\text{c2}} & \multirow{11}{*}{$56\times56$,197} & \blockbxc{64}{16}{4}  & \multirow{4}{*}{-} &  4$\times$4, 384, stride 4 & \multirow{4}{*}{$\times$1}& \multirow{11}{*}{$56\times56$,197} & \blockbxc{384}{96}{4}  & \multirow{4}{*}{-} &  4$\times$4, 384, stride 4 &\multirow{4}{*}{$\times$1} \\  \cline{5-5} \cline{10-10}
  & & & & \blocktransc{4} &  &  & & &  \blocktransbc{4} \\
  & & & & &  &  & & & &\\
  & & & & &  &  & & & &\\\cline{3-6}  \cline{8-11}
  & &\blockbc{64}{16}{4}  & \multirow{3}{*}{$[1\times1, 384]\longrightarrow$}  &  & \multirow{7}{*}{$\times$3} & &  \blockbc{384}{96}{4} & \multirow{3}{*}{$[1\times1, 576]\longrightarrow$} &   & \multirow{7}{*}{$\times$3} \\ 
    & & & & & & & & & &\\
     & & & & \blocktransc{4} & & & & & \blocktransbc{4}&\\
 &  & \ddashline& & & & & \ddashline& & &  \\ 
  & & \blockbc{64}{16}{4} &\multirow{3}{*}{$\longleftarrow[1\times1, 16]$}  &  & & & \blockbc{384}{96}{4} &\multirow{3}{*}{$\longleftarrow[1\times1, 96]$}  & & \\
  & & & & & & & & & &\\
  & & & & & & & & & & \\
\hline

\multirow{7}{*}{\text{c3}} &  \multirow{7}{*}{$28\times28$,197} & \blockbc{128}{32}{4} & \multirow{3}{*}{$[1\times1, 384]\longrightarrow$}  &  &\multirow{7}{*}{$\times$4} &  \multirow{7}{*}{$28\times28$,197} & \blockbc{768}{192}{4} & \multirow{3}{*}{$[1\times1, 576]\longrightarrow$}  & &\multirow{7}{*}{$\times$4} \\
 & & & & &  &  & & & &\\
 & & & & \blocktransc{4} & & & & & \blocktransbc{4}&\\
 &  & \ddashline& & & & & \ddashline& & &  \\ 
 & & \blockbc{128}{32}{4} &\multirow{3}{*}{$\longleftarrow[1\times1, 32]$}  &  & & & \blockbc{768}{192}{4} &\multirow{3}{*}{$\longleftarrow[1\times1, 192]$}  & & \\
 & & & & &  &  & & & &\\
 & & & & &  &  & & & &\\
\hline

\multirow{7}{*}{\text{c4}} &  \multirow{7}{*}{$14\times14$,197} & \blockbc{256}{64}{3} & \multirow{3}{*}{$[1\times1, 384]\longrightarrow$}  &  &\multirow{7}{*}{$\times$3} &  \multirow{7}{*}{$14\times14$,197} & \blockbc{1536}{384}{3} & \multirow{3}{*}{$[1\times1, 576]\longrightarrow$}  & &\multirow{7}{*}{$\times$3} \\
 & & & & &  &  & & & &\\
 & & & & \blocktransc{4} & & & & & \blocktransbc{4}&\\
 &  & \ddashline& & & & & \ddashline& & &  \\ 
 & & \blockbc{256}{64}{3} &\multirow{3}{*}{$\longleftarrow[1\times1, 64]$}  &  & & & \blockbc{1536}{384}{3}  &\multirow{3}{*}{$\longleftarrow[1\times1, 384]$}  & & \\
 & & & & &  &  & & & &\\
 & & & & &  &  & & & &\\
\hline

\multirow{7}{*}{\text{c5}} &  \multirow{7}{*}{$7\times7$,197} & \blockbc{256}{64}{3} & \multirow{3}{*}{$[1\times1, 384]\longrightarrow$}  &  &\multirow{7}{*}{$\times$1} &  \multirow{7}{*}{$7\times7$,197} & \blockbc{1536}{384}{3} & \multirow{3}{*}{$[1\times1, 576]\longrightarrow$}  & &\multirow{7}{*}{$\times$1} \\
 & & & & &  &  & & & &\\
 & & & & \blocktransc{4} & & & & & \blocktransbc{4}&\\
 &  & \ddashline& & & & & \ddashline& & &  \\ 
 & & \blockbc{256}{64}{3} &\multirow{3}{*}{$\longleftarrow[1\times1, 64]$}  &  & & & \blockbc{1536}{384}{3}  &\multirow{3}{*}{$\longleftarrow[1\times1, 384]$}  & & \\
 & & & & &  &  & & & &\\
 & & & & &  &  & & & &\\
\hline

\multicolumn{1}{c|}{\small Parameters} & \multicolumn{4}{c|}{\textbf{23.5 M}} & \multicolumn{6}{c}{\textbf{83.3 M}} \\
\hline
\multicolumn{1}{c|}{\small MACs} & \multicolumn{4}{c|}{\textbf{5.2 G}} & \multicolumn{6}{c}{\textbf{23.3 G}} \\ \bottomrule
\end{tabular}
}
\end{center}
\caption{Architecture of CNNTransformer-Ti and CNNTransformer-B, where MHSA-6/9 denotes the multi-head self-attention with heads 6/9 in transformer block and the fc layer is viewed as 1$\times$1 convolution here. And in output column, 56$\times$56,197 respectively mean the size of feature map is 56$\times$56 and the number of embedded patches is 197.}
\label{tab:arch_ti_b}
\vspace{-.5em}
\end{table*}
\renewcommand\arraystretch{1.0}

\end{document}